\documentclass[11pt]{article}

\usepackage[final]{acl}

\usepackage{times}
\usepackage{latexsym}

\usepackage[T1]{fontenc}

\usepackage[utf8]{inputenc}

\usepackage{microtype}

\usepackage{inconsolata}

\usepackage{graphicx}

\usepackage{comment}

\title{From Automation to Collaboration: Human-in-the-Loop Methods for Safe and Trustworthy NLP}

\author{
Most. Sharmin Sultana Samu$^{1}$\thanks{Corresponding author}, {\bf MD. Tanvir Ahmed Seum}$^{2}$, {\bf Md. Rakibul Islam}$^{1}$ \\[6pt]
$^{1}$Department of Computer Science and Engineering, BRAC University, Dhaka-1212. \\
$^{2}$Department of Electrical and Electronic Engineering, \\Rajshahi University of Engineering and Technology, Rajshahi-6204. \\[6pt]
\texttt{sharmin.sultana.samu@g.bracu.ac.bd}, 
\texttt{tanvirahmed3208@gmail.com}, \\
\texttt{md.rakibul.islam11@g.bracu.ac.bd}
}

\begin{document}
\maketitle
\vspace{5pt}
\begin{abstract}
Large language models are widely deployed in high-stakes NLP tasks, yet risks such as bias, hallucination, adversarial vulnerability and unreliable generalization remain. Probe-based auditing reveals inconsistencies in model behavior. Adversarial text generation uncovers robustness gaps, especially in lower-resourced languages with limited benchmarks. Enterprise text-to-SQL settings expose the difficulty of validating outputs over private and large-scale databases. Human supervision is essential for probe validation, adversarial verification and domain-specific annotation, but it is costly and hard to scale. This survey examines recent human-in-the-loop methods that shift NLP from automation toward collaboration for safety and trustworthiness. We review how human expertise supports auditing, robustness evaluation, data construction and model steering. Our findings highlight gaps in scalable probing, sustainable robustness benchmarks, low-resource settings and governance of private systems. We outline practical research directions for adaptive auditing, collaborative evaluation and accountable deployment.

\end{abstract}

\section{Introduction}
Natural Language Processing systems are commonly developed through a linear pipeline of training, fine-tuning and deployment. This workflow rarely incorporates structured user feedback after release. Model users often reveal overlooked cases and provide unseen data instances \cite{kreutzer2021offline}. Recent human-in-the-loop frameworks enable continuous feedback across development and deployment stages \cite{li2016dialogue, wallace2019trick}. Human intervention can occur during training through preference optimization \cite{stiennon2020learning} or during deployment through interactive correction \cite{hancock2019learning}. Large language models further raise concerns about bias, hallucination, adversarial vulnerability and unreliable generalization. Central research questions examine the role of human expertise, the integration of feedback into the NLP pipeline, the effect of feedback types on learning outcomes, bias mitigation and trust-aware system design.

The design space of human-in-the-loop NLP is broad and multidisciplinary. Feedback may originate from end users or crowd workers \cite{li2016dialogue, wallace2019trick}. Intervention may target data construction, model optimization, evaluation or post-deployment monitoring \cite{stiennon2020learning, hancock2019learning}. Effective systems must clearly communicate model needs, support intuitive interaction and learn reliably from feedback. Existing studies explore probe-based auditing, collaborative annotation and human-guided robustness evaluation. Cross-disciplinary synthesis across NLP and HCI remains limited.

\begin{figure*}[t] 
    \centering
    \includegraphics[width=0.98\textwidth]{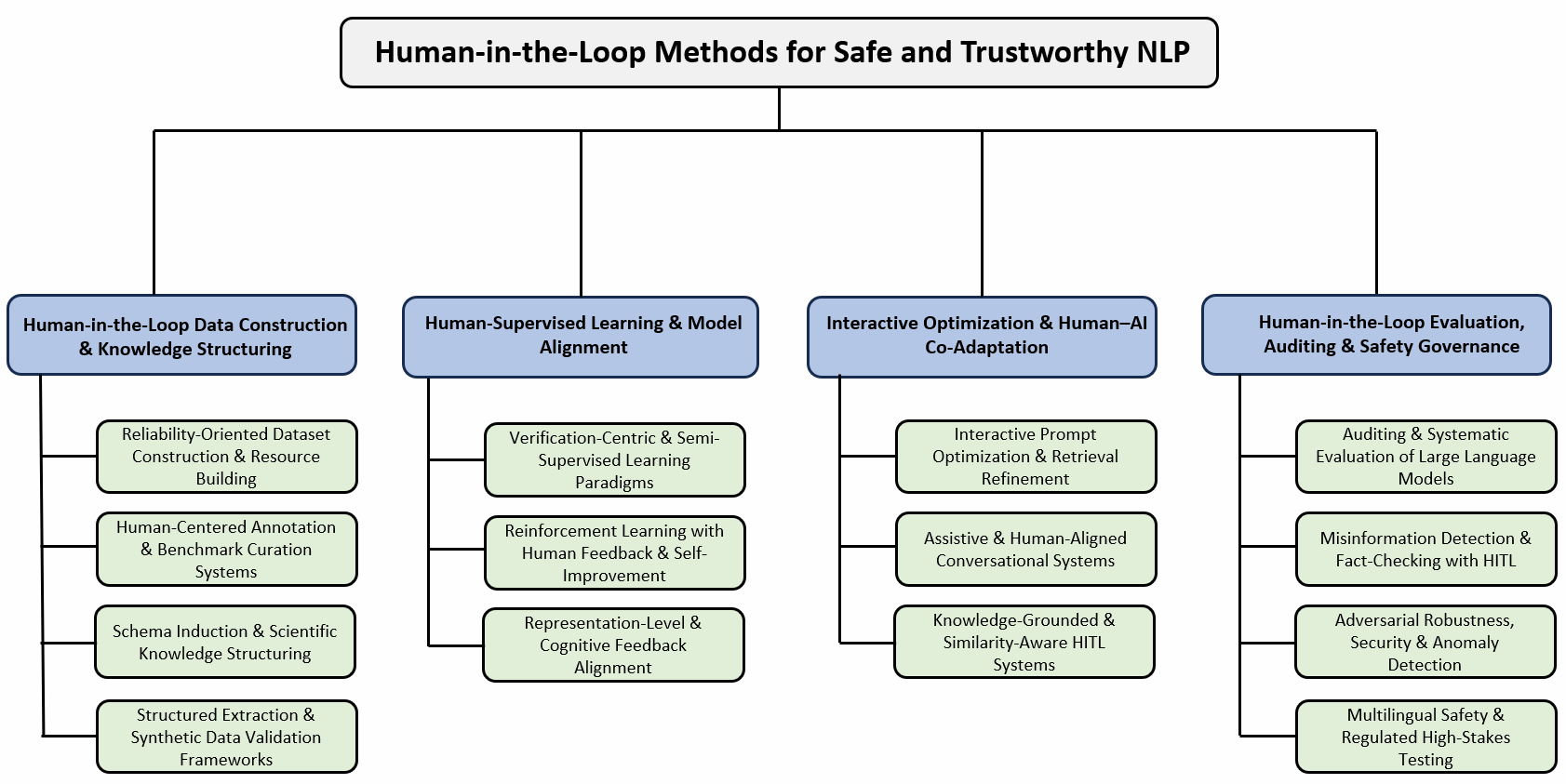} 
    \caption{Taxonomy of `Human-in-the-Loop Methods for Safe and Trustworthy NLP'.}
    \label{fig:taxonomy}
\end{figure*}

This survey presents a focused review of recent human-in-the-loop methods for safe and trustworthy NLP. The literature is organized into four taxonomies covering feedback sources, intervention stages, learning mechanisms and governance challenges. The taxonomy breakdown is presented in Figure~\ref{fig:taxonomy}. Each work is positioned by task setting, interaction design and feedback integration strategy. The review identifies gaps in scalability, low-resource adaptation, evaluation consistency and governance of deployed systems. It outlines future directions for adaptive auditing, collaborative robustness benchmarking, cost-aware supervision and accountable deployment.

\par
In short, with this study, we have made the following contributions:
\begin{enumerate}
    \item We provide synthesis of recent human-in-the-loop research for safe and trustworthy NLP.
    \item We organize the literature into four clear taxonomies.
    \item We identify key research gaps and outline concrete future directions toward collaborative and accountable NLP systems.
\end{enumerate}


\section{Human-in-the-Loop Data Construction \& Knowledge Structuring}
Current research examines how human feedback improves the quality and speed of tasks such as data labeling, disinformation detection and information extraction. Human-in-the-loop methods are vital when large, high-quality datasets are difficult to obtain. Human involvement offers a practical and cost-efficient training approach. It reduces reliance on fully manual work, which is slow and error-prone. It also addresses the limits of fully automated systems that struggle with nuance and domain-specific complexity. The integration of human expertise with machine learning produces NLP models that are more accurate, adaptable and reliable. Research covers applications such as scientific data curation, multilingual safety checks and event schema generation. These methods support both specialized and general NLP tasks.

    \subsection{Reliability-Oriented Dataset Construction \& Resource Building}
    \cite{bonet2023applying} and \cite{bonet2023semi} aim to reduce manual work through semi-automatic annotation. They differ in how they define reliability. \cite{bonet2023applying} introduces a three-phase HITL pipeline with active learning and pre-annotation. Humans review and correct machine outputs. The human role is explicit and measurable. An inter-annotator agreement of 0.70 shows annotation consistency. The method reduces time and achieves 95\% accuracy and F1-score on the RUN dataset. Lower performance for WHERE, WHY and HOW labels shows difficulty in complex semantic categories. \cite{bonet2023semi} applies multi-level semantic guidelines and linguistic analysis for reliability classification. It reports precision, recall, F1-score, accuracy and cost reduction. These metrics show efficiency and task performance. It does not report agreement statistics or detailed error analysis. This gap limits evaluation of annotation reliability. \cite{liu2025storm} emphasizes reasoning depth instead of annotation speed. It uses multiple LLM agents to generate samples. Expert mathematicians filter and validate outputs. Human scoring measures correctness and completeness. Very low baseline accuracy reflects dataset difficulty. Consistent fine-tuning improvements show effectiveness. The study does not report annotation efficiency or scalability. Resource cost remains unclear.

    \subsection{Human-Centered Annotation \& Benchmark Curation Systems}
    \cite{pangakis2025keeping}, \cite{schroeder2025just} and \cite{wenz2025benchpress} emphasize human oversight. \cite{thielmann2024human} reduces direct human effort through few-shot supervision and topic extraction. \cite{pangakis2025keeping} uses iterative prompt refinement with GPT-4 and repeated sampling to measure consistency. The median accuracy and F1 is 0.850 and 0.707 respectively. Nine tasks report precision or recall below 0.5. These results show instability in automated annotation. BERT outperforms GPT-4 with larger labeled datasets. This outcome limits claims about prompt-based scalability. \cite{schroeder2025just} applies a controlled experiment to examine AI support in crowd annotation. Krippendorff’s alpha defines baseline agreement. LLM suggestions increase annotator confidence and label overlap. Annotation time does not decrease. Higher F1 scores on AI-reviewed labels indicate anchoring bias. Benchmark validity is affected. \cite{wenz2025benchpress} combines retrieval-augmented generation with expert validation. It evaluates accuracy, latency and semantic fidelity through backtranslation. Accuracy reaches 93\%. Latency decreases. Rubric-based semantic checks improve benchmark reliability. Resource cost and expert dependence are not specified. \cite{thielmann2024human} shows that one labeled document per class can produce coherent topics. NPMI exceeds unsupervised baselines. Classification accuracy indicates structural alignment. Human validation steps are not described. Annotation robustness remains uncertain. Dimensionality reduction is not applied. Scalability to noisy real-world benchmarks is not tested.

    \subsection{Schema Induction \& Scientific Knowledge Structuring}
     \cite{zhang2023human}, \cite{john2026extractable} and \cite{sadruddin2025llms4schemadiscovery} treat humans as central to schema induction. They implement collaboration at different levels. \cite{zhang2023human} focuses on interactive graph construction and ontology grounding. It evaluates accuracy, graph edit distance and grounding success. High generation accuracy contrasts with many structural edits. Language fluency does not ensure reasoning reliability. Human correction effort is visible. \cite{john2026extractable} emphasizes workflow usability. It reports a System Usability Scale score of 84.17 and strong time reduction. Efficiency gains are clear. Effectiveness partly depends on participant perception. Structural validation is weaker than edit distance or similarity metrics. \cite{sadruddin2025llms4schemadiscovery} applies staged validation on curated and large corpora. It evaluates schemas with ROUGE, BLEU and BERTScore. These metrics provide standardized semantic comparison. Human correction effort is not directly measured. Model variability and redundancy indicate instability in automated schema generation. Human validation improves accuracy in all three studies. \cite{zhang2023human} highlights correction cost, \cite{john2026extractable} prioritizes speed and usability. \cite{sadruddin2025llms4schemadiscovery} prioritizes metric-based semantic coherence.
    
    \subsection{Structured Extraction \& Synthetic Data Validation Frameworks}
    \cite{kang2024human} and \cite{chhetri2025structsense} treat structured extraction as a collaborative inspection process. They differ in reliability definition and measurement. \cite{kang2024human} places human selection within an interface that shows transformation and feature provenance. Quality is measured with label alignment, grammaticality and perplexity scores normalized between 0 and 1. Efficiency is assessed by the number of correctly identified texts. Robustness is evaluated with DeepWord attack success rate. Participants detect three to four times more correct texts. Robustness increases by up to 32\%. Labeling accuracy shows no statistically significant improvement. Interface support increases confidence and efficiency more than correctness. \cite{chhetri2025structsense} organizes extraction into extractor, alignment, judge and feedback agents guided by domain ontologies. Performance is measured with precision, recall, F1 score, ontology alignment rate, judge accuracy and Shannon diversity index. Human feedback improves recall and F1, especially for weaker models. Quality gains extend beyond perception-based metrics used in \cite{kang2024human}. Task variability and smaller gains for strong baselines show limits of iterative refinement. \cite{kang2024human} highlights inspection effort and user confidence but does not improve core accuracy. \cite{chhetri2025structsense} achieves measurable extraction gains through ontology grounding and feedback loops. User workload and interaction cost remain less explicit.

    \subsection{Datasets for Human-in-the-Loop Data Construction \& Knowledge Structuring}
    \cite{bonet2023applying} uses a small curated RUN dataset of Spanish news with detailed 5W1H annotations. \cite{bonet2023semi} relies on similar news data for semantic labeling and disinformation detection, with emphasis on annotation quality. \cite{liu2025storm} builds a synthetic mathematical derivation dataset filtered by human experts to ensure reasoning complexity. \cite{thielmann2024human} and \cite{pangakis2025keeping} use benchmark and manually annotated corpora for topic modeling and social science tasks. \cite{zhang2023human}, \cite{john2026extractable} and \cite{sadruddin2025llms4schemadiscovery} create human-curated schema datasets from GPT outputs, ORKG comparisons and scientific papers. These datasets support domain coverage and iterative validation. \cite{schroeder2025just} and \cite{wenz2025benchpress} integrate community conversations or enterprise SQL logs with AI-assisted labeling to test LLMs on real-world data.


\section{Human-Supervised Learning \& Model Alignment}
Human-supervised learning and model alignment aim to keep NLP systems reliable during updates and deployment. Core challenges include lowering annotation cost and improving quality with limited human input. Research evaluates manual verification as an alternative to full annotation. Studies address annotation and verification errors. Semi-supervised workflows reduce labeling effort and limit error amplification. Interactive learning uses ranking and exemplar-based feedback instead of binary labels. Alignment research analyzes reinforcement learning with human feedback and identifies risks such as reward poisoning. Adaptive test-time learning applies reflective reasoning and human guidance. Explainable moderation detects harmful content with transparent justification. Representation-level control aligns internal behavior with human values. Recent work studies implicit preference signals and multilingual continual adaptation without performance loss.

    \subsection{Verification-Centric \& Semi-Supervised Learning Paradigms}
    \cite{weber2021better} presents a verification-focused semi-supervised pipeline that accepts correct model predictions and sends only errors for full annotation. It integrates self-training and Active Learning, and simulates iterative updates in a production NLU system. Evaluation uses task-level metrics such as Semantic Error Rate, Intent Classification Error Rate, F-SemER, F-ICER, entropy, annotation volume and cost reduction. Results show up to 97\% lower annotation, nearly 60\% cost savings, improved recall and F1, and 66\% fewer human errors, with larger gains for BERT. \cite{martin2026beyond} replaces binary labels with probabilistic ranking and exemplar queries guided by Bayesian inference and cost-aware selection. Human feedback is modeled in embedding space to optimize information gain under annotation cost. Metrics include sample complexity, mean squared error, classification accuracy, information rate and response time. It achieves up to 85\% fewer human interactions, over 57\% time savings and ranking queries reduce learning error more than label-based feedback. \cite{weber2021better} shows strong practical impact but depends on prediction quality and verification. \cite{martin2026beyond} provides a principled framework for richer supervision but adds computational complexity via variational approximations and probabilistic modeling.
        
    \subsection{Reinforcement Learning with Human Feedback \& Self-Improvement}
    \cite{wang2024rlhfpoison} questions RLHF reliability by showing preference rankings can be poisoned. \cite{samani2025large} uses RLHF and Direct Policy Optimization to improve alignment and explainable sexism detection. \cite{he2025enabling} emphasizes reflective self-improvement via expert queries during deployment. \cite{samani2025large} and \cite{he2025enabling} assume human feedback improves behavior. \cite{wang2024rlhfpoison} shows feedback can be exploited with little visible safety impact. Evaluation reflects these goals. \cite{wang2024rlhfpoison} measures malicious goal achievement, safety accuracy, harmfulness ratio and output length shifts. \cite{samani2025large} uses F1-Macro to track classification gains after fine-tuning and RLHF. \cite{he2025enabling} evaluates sensitivity, specificity, task accuracy and handling time. \cite{samani2025large} reports stable accuracy improvements, especially for large models. \cite{he2025enabling} shows iterative guidance increases accuracy and reduces processing time. \cite{wang2024rlhfpoison} reveals alignment metrics can remain stable even when behavior covertly shifts toward harmful patterns.

    \subsection{Representation-Level \& Cognitive Feedback Alignment}
    \cite{liu2024aligning} aligns models by manipulating internal activity through Representation Alignment from Human Feedback. \cite{harada2025cognitive} derives preferences from EEG signals and optimizes behavior via Direct Preference Optimization without an explicit reward model. \cite{liu2024aligning} uses contrastive instruction tuning, difference vectors and Low-Rank Adapters at the representation level. \cite{harada2025cognitive} builds a cognitive decoder to translate neural signals into preference pairs. Evaluation differs by supervision source. \cite{liu2024aligning} measures benchmark accuracy, win rates, MT-Bench scores, toxicity detection and human judgments. \cite{harada2025cognitive} measures adjusted win rate, sentiment classification and decoder accuracy. Results show \cite{liu2024aligning} improves benchmark performance, human-rated alignment and shifts internal representations toward preferred directions. \cite{harada2025cognitive} shows cognitive signals can approximate human and AI feedback, increasing win rates and positivity as training pairs grow. \cite{liu2024aligning} offers stable internal control and broad task gains but depends on identifiable preference directions. \cite{harada2025cognitive} reduces manual annotation but needs specialized EEG data and accurate decoding.

    \subsection{Datasets for Human-Supervised Learning \& Model Alignment}
    \cite{weber2021better} uses millions of user utterances with expert-validated test sets and Active Learning samples, grounding alignment in production-scale interactions. \cite{martin2026beyond} relies on crowdsourced sentiment and image aesthetics datasets with recorded response times for controlled study of rich queries in simulated settings. \cite{wang2024rlhfpoison} analyzes preference-based alignment and poisoning risks using PKU-SafeRLHF and Stanford Alpaca. \cite{he2025enabling} tests adaptive learning on regulatory screening and contract datasets with simulated concept drift. \cite{samani2025large} employs the EDOS benchmark for hierarchical sexism detection under class imbalance, focusing on explainable moderation. \cite{liu2024aligning} uses UltraFeedback and Anthropic Helpful-Harmless dialogue data to extract preference-linked internal representations. \cite{harada2025cognitive} applies a small EEG-based sentiment corpus with limited participants, introducing cognitive signals with narrow scope.

\section{Interactive Optimization \& Human–AI Co-Adaptation}
As large language models enter complex real-world settings, challenges shift from isolated prediction to interactive optimization and co-adaptation. Problem domains include retrieval-augmented question answering, interactive prompt refinement, collaborative discourse analysis, procedural guidance, semantic intent clustering, similarity discovery across heterogeneous data, dictionary-based content analysis, knowledge graph QA and test-time adaptation. These tasks reveal limits of static automation, unreliable evaluation, shallow embeddings, rigid pipelines, generative hallucinations and lack of continuous learning. Key research questions explore how human feedback can guide retriever tuning, prompt selection, discourse interpretation, clustering, similarity assessment and structured querying. Other work studies how AI guidance affects human performance and learning. Some studies examine agent reflection on uncertainty, targeted human input and knowledge updating during deployment. Current approaches emphasize iterative feedback, transparent evaluation and shared human-model control for reliable adaptation.

    \subsection{Interactive Prompt Optimization \& Retrieval Refinement}
    \cite{afzal2024towards} studies full retrieval-augmented generation pipelines using dense and vector retrievers with GPT-4 and fine-tuned LongT5. Performance drops with noisy or variable HR data, and query transformation adds little. Automatic metrics often misalign with human judgments that show limits of reference-based evaluation. \cite{li2025iprop} focuses on iterative prompt refinement, letting users guide paraphrasing and adjustments. F1 scores improve steadily that shows human guidance enhances task performance, though it depends on user expertise and is not fully automated. Compared to \cite{afzal2024towards}, \cite{li2025iprop} gives more prompt-level control, while \cite{afzal2024towards} addresses end-to-end generation with broader complexity. Both show trade-offs: system pipelines handle diverse tasks but face evaluation issues, while prompt optimization yields precise gains but relies on human effort.
    
    \subsection{Assistive \& Human-Aligned Conversational Systems}
    \cite{cohn2024towards} shows chain-of-thought prompting with active learning lets models capture nuanced discourse near human level, though GPT-4 still hallucinates or misses subtle details. \cite{bellos2025towards} finds real-time AI guidance in physical tasks improves success and reduces errors while helping users learn, though instruction relevance can vary by task. \cite{hong2025dial} demonstrates that combining language models with iterative clustering enhances intent discovery in dialogues, producing clusters close to human judgments and improving classification, but requires structured roles and careful refinement. All approaches use human-in-the-loop feedback, with trade-offs: textual discourse gains nuanced alignment, procedural tasks gain interactive guidance and dialogue systems benefit from structured evaluation. Balancing interpretability, domain adaptability and real-world applicability remains key for assistive conversational AI.

    \subsection{Knowledge-Grounded \& Similarity-Aware HITL Systems}
    \cite{zeng2024similar} applies multi-stage summarization and hidden-state similarity to help experts identify patterns in image and tabular data, providing intuitive insights but struggling with outliers. \cite{bakhteev2025embed2discover} targets structured text classification with embeddings, active learning and interpretable models, achieving high recall and reducing annotation time, but is limited to parliamentary datasets. \cite{pusch2026human} uses interactive query generation over knowledge graphs, letting users refine LLM-generated Cypher queries while maintaining accuracy, though performance varies by domain. \cite{zeng2024similar} excels in pattern recognition, \cite{bakhteev2025embed2discover} in efficient interpretable text annotation and \cite{pusch2026human} in guiding human-AI reasoning for structured queries.

    \subsection{Datasets for Interactive Optimization \& Human–AI Co-Adaptation}
    HR chatbots use curated FAQ pairs and real user utterances with metadata for structured retrieval and QA triplets \cite{afzal2024towards}. Emotion and discourse studies use small, annotated corpora from classroom interactions and student logs for fine-grained evaluation of human-aligned reasoning \cite{cohn2024towards, li2025iprop}. Multimodal and procedural datasets combine video, audio and annotations to capture complex human-AI interactions \cite{bellos2025towards, zeng2024similar}. Large-scale dialogue and document collections support intent discovery, clustering and legislative text analysis, often with noisy inputs and expert labeling \cite{hong2025dial, bakhteev2025embed2discover}. Knowledge graph experiments use synthetic and domain-specific graphs to study explainability and interactive querying \cite{pusch2026human}.

\section{Human-in-the-Loop Evaluation, Auditing \& Safety Governance}
As NLP shifts from automation to collaborative intelligence, evaluation, auditing and safety governance become critical. Research addresses bias, hallucination, inconsistency and hidden failure patterns in large language models. Structured auditing frameworks combine automated analysis with human judgment to detect visible and subtle weaknesses. Human-validated probes and interactive evaluation enhance transparency and reliability. Multilingual safety gaps and low-resource language vulnerabilities gain attention. Robustness benchmarking and adversarial testing require continuous oversight. Misinformation detection, phishing prevention and scalable fact-checking illustrate governance challenges online. Early identification of harmful content relies on realistic, human-grounded evaluation. Human-in-the-loop grading and data extraction improve reliability in education and evidence synthesis. In predictive tasks, human correction reduces misplaced trust. Safety-critical domains, like air traffic control, demand expert-driven evaluation aligned with operational standards.

\subsection{Auditing \& Systematic Evaluation of Large Language Models}
\cite{bonet2023applying} and \cite{amirizaniani2025developing} use two-phase auditing where one model generates probes and another answers them, that reduces self-evaluation bias. Iterative prompt refinement with a human-guided Annotator Codebook improves probe relevance but increases cost and limits scalability. Evaluation includes BERTScore, ROUGE-L, BLEURT, GPT\_Judge, Bias Score, Stereotype Score, Idealized CAT Score and Accuracy on benchmarks like TruthfulQA, HaluEval2.0, BBQ and StereoSet. Cohen’s Kappa, Krippendorff’s Alpha and overlap rate track agreement. Both outperform Vanilla and DSPy baselines, reveal failure cases in Falcon 7B, Llama 2-7B and GPT-3.5-turbo, and match Adatest++ probe quality, but rely on metrics and benchmark coverage, missing deeper reasoning flaws. \cite{chu2025llm} uses human-in-the-loop grading with chain-of-thought prompting, rubric inquiry and RL optimization, improving consistency but requiring expert involvement. \cite{schroeder2025large} applies staged LLM data extraction with human agreement, showing strong results for explicit data but weaker for categorized information, using simpler metrics and realistic prompting without structured probes or bias auditing.

\subsection{Misinformation Detection \& Fact-Checking with HITL}
\cite{yang2021scalable} organizes large volumes of overlapping content using semantic clustering and summarization. Sentence-Transformers with Agglomerative Clustering and Leiden community detection group short messages, followed by extractive and abstractive summarization evaluated with ROUGE, BERTScore, Silhouette coefficient and expert ratings. Leiden clustering and extractive summarization perform best, with human scores above 4. The framework reduces 28,818 tweets to 700 claims, showing scalability, but focuses on consolidation rather than early risk detection. \cite{mendes2023human} builds an end-to-end pipeline with claim extraction, trend analysis via Fisher’s Exact Test, stance classification and human moderation. Evaluation measures relevance, timeliness, policy accuracy and analyst workload. It detects 50\% of novel misleading COVID-19 treatment claims with a 21-day median lead time and identifies 124.2 policy violations per hour, though stance detection F1 is 66.7. \cite{yang2021scalable} excels in semantic organization and content clarity, while \cite{mendes2023human} excels in early intervention and operational effectiveness.

\subsection{Adversarial Robustness, Security \& Anomaly Detection}
\cite{asiri2024towards, cao2025human, deng2024reliable} implement human oversight at different robustness and security layers. \cite{asiri2024towards} embeds feedback in a live phishing detection loop with a browser extension and isolation server, retrains a deep model on uncertainty and evaluates Accuracy, Precision, Recall and F1. Accuracy rises from 0.75 to 0.80 with ~5\% gains in other metrics, though evaluation lacks fine-grained error analysis. \cite{cao2025human} uses HITL-GAT for iterative attack generation, human validation and benchmarking with Edit Distance, Cosine Similarity, BERTScore and human ratings, reporting robustness scores of 0.56–0.57 for CINO variants. The benchmark fills a low-resource gap and emphasizes perceptual quality, but averages may mask vulnerabilities. \cite{deng2024reliable} focuses on failure diagnosis using attribution-aware clustering, UMAP, K-means, human annotation, label propagation and Core Risk Minimization with Gaussian noise. It improves F1 by 18.49 ± 2.7\%, reduces annotation time to 8.3 ± 2.1 minutes and aligns attention with true abnormal segments. \cite{asiri2024towards} emphasizes continual adaptation, \cite{cao2025human} emphasizes benchmark realism and \cite{deng2024reliable} emphasizes interpretable correction and scalable propagation. \cite{asiri2024towards} shows modest transparency gains, \cite{cao2025human} improves robustness assessment with strong human filtering and \cite{deng2024reliable} achieves the largest performance gains with explicit attention alignment and efficient human effort.

\subsection{Multilingual Safety \& Regulated High-Stakes Testing}
\cite{chualost} and \cite{carvell2026human} focus on safety under different pressures with human judgment central to evaluation. \cite{chualost} builds a multilingual benchmark using Generate–Label–Translate with LLM red teaming, majority-voted LLM annotation and expert verification within a structured harm taxonomy. Evaluation includes inter-annotator agreement of 0.70–0.80, Cohen’s kappa of 0.68–0.72, F1 for unsafe content and cosine similarity for translation fidelity. F1 drops over 30\% on non-English data, with severe degradation in Tamil, revealing weaknesses in guardrails, keyword reliance and shallow intent modeling. GPT-4O mini preserves semantics in Chinese and Malay, but downstream classifiers struggle. \cite{carvell2026human} evaluates AI agents in air traffic control via Project Bluebird, with human instructors grading competency using Intraclass Correlation, Spearman’s rho and trajectory deviation thresholds. Rule-based agents control well, optimization agents coordinate better and iterative instructor feedback improves performance, yet all fall below trainee standards, especially in safety. \cite{chualost} highlights multilingual blind spots, while \cite{carvell2026human} reveals operational gaps. One emphasizes scalability and cross-lingual coverage, the other realism and accountability.

\subsection{Datasets for Human-in-the-Loop Evaluation, Auditing \& Safety Governance}
\cite{amirizaniani2024llmauditor} and \cite{amirizaniani2025developing} audit TruthfulQA, expanding to HaluEval2.0, BBQ and StereoSet for hallucination and social bias, though probes cover small subsets. \cite{chu2025llm} uses 1,376 expert-graded short answers, enabling controlled evaluation but limited domain breadth. \cite{schroeder2025large} tests LLM-assisted data extraction on 8- and 112-study review datasets with 24 variables per study, supporting realistic workflows but modest scale. \cite{yang2021scalable} processes 28,818 tweets linked to 959 articles, focusing on redundancy reduction. \cite{mendes2023human} uses 14.7 million COVID-19 tweets with crowd-annotated stance labels and policy review, providing large-scale realism in a concentrated domain. \cite{asiri2024towards} compiles 50,000 phishing and benign URLs with synthetic generation, ensuring class balance but risking artifacts. \cite{cao2025human} benchmarks Tibetan TNCC-title and TU\_SA datasets for low-resource adversarial testing, offering linguistic diversity with limited scope. \cite{deng2024reliable} evaluates anomaly detection on Mars rover, ECLSS and industrial KPI time series with labeled abnormal segments for high-stakes validation. \cite{chualost} builds a 5,364-example multilingual safety benchmark across Singlish, Chinese, Malay and Tamil with toxicity-preserving translations. \cite{carvell2026human} relies on proprietary air traffic control logs, simulator traces and regulatory documents, ensuring operational realism but limiting reproducibility.

\section{Key Findings and Trending Research Directions}
Human-in-the-loop NLP research faces persistent challenges in human dependence, limited domain coverage and incomplete automation. Small datasets restrict reliable training and validation. Expert-driven selection and annotation increase cost and reduce scalability. Bias, anchoring effects and language imbalance affect feedback quality and dataset construction. Proprietary data and private enterprise settings limit reproducibility and open benchmarking. RLHF frameworks show structural fragility and high computational demand. Cognitive feedback studies rely on small controlled datasets and narrow task settings. Prompt sensitivity, clustering instability and similarity matrix scalability weaken robustness. Privacy risks and feedback error propagation raise governance concerns.

Trending research directions focus on reducing human workload while preserving reliability. Larger and more diverse datasets are required across multilingual and multi-domain settings. Stronger automated pre-annotation, extraction and ensemble validation methods are needed. Domain-adaptive workflows and robust evaluation metrics must ensure semantic alignment in collaborative datasets. Scalable human validation protocols, fair annotator governance and privacy-aware feedback integration are essential. Open benchmarking, cross-domain evaluation and regulator engagement can strengthen accountability. Advancing from automation to collaboration demands resilient, scalable and governance-aware human-AI partnership in NLP.

\section{Conclusion}
Human-in-the-loop NLP addresses the limits of fully automated pipelines in safety-critical and user-facing applications. Systems aim to improve performance while supporting interpretability and usability. Designs vary across feedback sources, intervention stages and interaction types. Many rely on crowd workers and focus mainly on NLP. Challenges remain in scalable feedback integration, interface design, bias control and model governance. Weak interfaces risk inconsistent or misleading feedback. Limited user studies constrain understanding of real-world effectiveness. Gaps exist in cross-disciplinary collaboration, evaluation protocols, interface rigor and feedback transparency. Deeper HCI involvement, interactive visualization, systematic user studies and open feedback sharing are needed. Progress requires structured human-AI partnerships for safe and trustworthy NLP.

\section*{Limitations}
This survey has several limitations. The field is evolving rapidly, especially with large language models and agentic systems, which may introduce new paradigms beyond our coverage. The surveyed studies span diverse tasks such as auditing, annotation, adversarial generation, alignment and benchmarking, which makes direct comparison difficult due to inconsistent metrics and evaluation settings. Many works are recent and lack long-term validation in real-world deployments. The proposed four taxonomies reflect our synthesis of current trends and design choices. Alternative categorizations may reveal different perspectives on collaboration, governance and feedback integration.

\section*{Acknowledgments}
This study was conducted collaboratively to review human-in-the-Loop method for safe and trustworthy NLP, with each author contributing equally.


\bibliography{custom}

\appendix
\section{Paper Selection Criteria}
To ensure a comprehensive and high-quality survey on human-in-the-loop approaches for safe and trustworthy NLP, we adopted a structured paper selection strategy. The selection process focused on identifying recent and influential studies that integrate human feedback, collaboration, validation, supervision, or intervention within NLP systems, particularly in the context of trustworthiness, safety, explainability, alignment, evaluation, and data quality.

We primarily considered papers published between 2021 and 2026, reflecting the rapid evolution of large language models, reinforcement learning from human feedback, human-centered AI, and trustworthy NLP systems. The literature collection included publications from major NLP and AI venues such as ACL, EMNLP, NAACL, COLING, LREC-COLING, SIGIR, and related workshops, as well as high-quality journals and recent arXiv preprints addressing emerging research directions.

The paper identification process was conducted using keyword-based searches over digital libraries including the ACL Anthology, IEEE Xplore, ACM Digital Library, SpringerLink, ScienceDirect, and arXiv. Search keywords included combinations of terms such as human-in-the-loop NLP, trustworthy AI, LLM auditing, human feedback, RLHF, interactive NLP, human-centered AI, safe NLP systems, LLM alignment, annotation with humans, and human-guided evaluation.

The final corpus of selected papers was curated through iterative screening based on title, abstract, methodology, and contribution relevance. Studies were retained if they substantially incorporated human participation into the NLP pipeline or addressed safety, trustworthiness, reliability, interpretability, or alignment challenges through human collaboration mechanisms.

\subsection{Inclusion Criteria}

The following inclusion criteria were applied during the paper selection process:

\begin{itemize}
    \item \textbf{Human-in-the-Loop Focus:} Papers must explicitly incorporate human feedback, human validation, human supervision, interactive learning, or collaborative decision-making mechanisms within NLP or LLM systems.

    \item \textbf{Relevance to Safe and Trustworthy NLP:} Studies addressing trustworthiness dimensions such as safety, fairness, explainability, reliability, robustness, auditing, alignment, misinformation detection, or responsible AI were included.

    \item \textbf{NLP or LLM-Centered Research:} Only papers directly related to NLP tasks, language models, text generation, language understanding, conversational AI, information extraction, or text-centric AI systems were considered.

    \item \textbf{Peer-Reviewed or Emerging Influential Work:} Publications from recognized conferences, journals, workshops, or influential recent arXiv preprints were included to capture both mature and emerging research trends.

    \item \textbf{Recent Publications (2021--2026):} The review emphasized contemporary developments corresponding to the rise of LLMs and modern human-centered AI methodologies.

    \item \textbf{Methodological Contribution:} Papers proposing frameworks, architectures, workflows, datasets, evaluation pipelines, annotation strategies, or benchmarking methodologies involving HITL components were prioritized.
\end{itemize}

\subsection{Exclusion Criteria}

The following exclusion rules were applied to refine the survey scope:

\begin{itemize}
    \item \textbf{Non-NLP HITL Research:} Studies focusing exclusively on computer vision, robotics, healthcare monitoring, cybersecurity, or general machine learning without a significant NLP component were excluded.

    \item \textbf{Purely Automated Systems:} Papers that did not involve any meaningful human participation, feedback, or intervention mechanism were omitted.

    \item \textbf{Irrelevant Trustworthiness Contexts:} Studies unrelated to safety, trustworthiness, alignment, interpretability, auditing, or human-centered evaluation were excluded even if they mentioned HITL superficially.

    \item \textbf{Duplicate or Redundant Publications:} Earlier versions of substantially similar works, duplicate preprints, or overlapping studies were removed to avoid redundancy.

    \item \textbf{Insufficient Technical Detail:} Papers lacking methodological clarity, experimental validation, or substantive discussion of HITL mechanisms were excluded.

    \item \textbf{Non-English or Inaccessible Publications:} Articles without accessible full texts or not written in English were not considered in the final review set.
\end{itemize}

\subsection{Screening and Refinement Process}

The collected papers were screened in three stages:
(i) title-level filtering,
(ii) abstract and contribution analysis, and
(iii) full-text examination.

Priority was given to works that demonstrated practical integration of humans within the NLP lifecycle, including data annotation, model alignment, evaluation, auditing, adversarial testing, prompt optimization, and collaborative reasoning. Additionally, papers from high-impact venues such as ACL-related conferences and reputable publishers including Elsevier, ACM, and IEEE were given higher priority due to their research quality and community impact.

\section{Limitations and Future Works} \label{sec:appendix}
In this section, we discuss the limitations and future research directions associated with each taxonomy.

\subsection{Limitations \& Future Directions in Human-in-the-Loop Data Construction and Knowledge Structuring}
Limitations and future directions focus on human dependence, domain coverage and automation gaps. Small datasets, such as in \cite{bonet2023applying}, restrict training and validation. Bias and anchoring effects appear in \cite{bonet2023applying} and \cite{schroeder2025just}. Human expertise is essential for high-quality selection in \cite{liu2025storm}, \cite{zhang2023human} and \cite{sadruddin2025llms4schemadiscovery}. Expert involvement raises cost and limits scalability. Language and cultural bias are reported in \cite{zhang2023human, kang2024human, pangakis2025keeping}. Tools are sometimes outdated or limited, as shown in \cite{bonet2023applying} and \cite{kang2024human}. Future research calls for larger and more diverse datasets. Studies propose stronger automated pre-annotation and extraction models. Many aim to reduce human workload through iterative and ensemble methods. Expansion to multilingual, multi-domain and enterprise settings is required. \cite{john2026extractable} and \cite{sadruddin2025llms4schemadiscovery} stress domain-adaptive workflows and robust evaluation metrics to ensure reliability and semantic alignment in collaborative NLP datasets.

\subsection{Limitations \& Open Challenges in Human-Supervised Learning \& Model Alignment}
Scalability and robustness are key concerns across these studies. \cite{weber2021better} and \cite{he2025enabling} rely on sustained human involvement. \cite{weber2021better} assumes a mature system with reliable verification. \cite{he2025enabling} depends on continuous expert guidance and expanding knowledge, limiting large-scale deployment. \cite{martin2026beyond} and \cite{liu2024aligning} use principled modeling frameworks, but computational costs and architecture limit high-dimensional or large-model feasibility. \cite{wang2024rlhfpoison} reveals RLHF structural fragility under white-box assumptions. \cite{samani2025large} shows aligned RLHF struggles with fine-grained tasks and real-time moderation due to resource demands. \cite{harada2025cognitive} applies cognitive signals, but small EEG datasets and pairwise feedback limit generalization beyond controlled sentiment settings.

\subsection{Limitations \& Future Directions in Interactive Optimization \& Human–AI Co-Adaptation}
 Limitations stem from dataset scale, domain specificity and accessibility. Proprietary models and private HR data restrict sharing and replication \cite{afzal2024towards}. Small samples and single-annotator evaluations reduce statistical power and may introduce bias \cite{cohn2024towards, li2025iprop}. Multimodal and LLM-based methods face challenges in interpretability and domain generalization \cite{zeng2024similar, bellos2025towards}. Single-language or domain-specific datasets limit broader applicability \cite{hong2025dial, bakhteev2025embed2discover}. Synthetic or limited query datasets may not capture real-world complexity \cite{pusch2026human}. Future work should expand dataset diversity and sample sizes, support multilingual and cross-domain evaluation, integrate richer human feedback, and develop scalable annotation, robust evaluation and validation methods for interactive HITL systems.
 
\subsection{Limitations \& Governance Challenges in Human-in-the-Loop Evaluation, Auditing \& Safety Governance}
\cite{amirizaniani2024llmauditor} and \cite{amirizaniani2025developing} risk high evaluation cost, annotator marginalization and focus on QA with researcher-led assessment. \cite{chu2025llm} evaluates one LLM and ignores discarded responses, limiting generalizability. \cite{schroeder2025large} shows prompt sensitivity, hallucination and PDF constraints, weakening reliability in reviews. \cite{yang2021scalable} lacks gold clustering labels, oracle summaries and faces similarity matrix scalability issues. \cite{mendes2023human} depends on domain-specific extractors, mock moderators and unstable policies, raising bias and deployment concerns. \cite{asiri2024towards} has privacy risks, feedback error propagation and limited phishing diversity. \cite{cao2025human} is confined to one low-resource language and needs evolving adversarial strategies. \cite{deng2024reliable} faces clustering errors and limited human validation scalability. \cite{chualost} has taxonomy inconsistencies, potential LLM bias and regional limits. \cite{carvell2026human} excludes communication skills and limits replication. Future directions emphasize broader datasets, cross-domain evaluation, scalable human validation, fair annotator governance, privacy protection, open benchmarking, improved simulation and regulator engagement.

\section{Dataset Summary Across Studies}

Table \ref{tab:dataset_summary} summarizes the datasets used in the reviewed human-in-the-loop studies for safe and trustworthy NLP. The datasets span diverse domains including misinformation detection, LLM alignment, auditing, dialogue systems, multilingual safety, knowledge graphs, and human-centered evaluation, highlighting the broad applicability of HITL methodologies across NLP tasks.

\begin{table*}[t]
\centering
\scriptsize
\resizebox{\textwidth}{!}{
\begin{tabular}{p{2.5cm} p{3.2cm} p{8.0cm} p{2.8cm}}
\hline
\textbf{Reference} & \textbf{Dataset(s)} & \textbf{Summary} & \textbf{Domain/Task} \\
\hline

\cite{bonet2023applying} & RUN Dataset & Contains 170 Spanish news items evenly divided between reliable and unreliable content with fine-grained 5W1H annotations using the RUN-AS guideline. & Fake news detection \\

\cite{amirizaniani2025developing} & TruthfulQA, HaluEval2.0, BBQ, StereoSet & Benchmark datasets for hallucination and social bias auditing of LLMs. & LLM auditing \\

\cite{amirizaniani2024llmauditor} & TruthfulQA & Factual QA benchmark across 38 categories used for hallucination detection and probe validation. & Hallucination detection \\

\cite{de2024human} & CIFAR-10 & Image dataset with 60,000 images across 10 object categories. & Federated learning \\

\cite{weber2021better} & Dialogue Utterance Dataset & Millions of unlabeled utterances with expert annotations and active learning samples. & Natural language understanding \\

\cite{yang2021scalable} & MM-COVID Dataset & COVID-19 misinformation dataset containing tweets and news articles. & Misinformation detection \\

\cite{cao2025human} & TNCC-title, TU\_SA & Tibetan datasets for news classification and sentiment analysis. & Low-resource NLP \\

\cite{kang2024human} & SST-2, TweetEval & Sentiment and hate speech benchmark datasets with synthetic text augmentation. & Text classification \\

\cite{mendes2023human} & COVID-19 Twitter Dataset & Large-scale Twitter dataset with 14.7M tweets related to COVID-19 treatments. & Early misinformation detection \\

\cite{john2026extractable} & ORKG Experimental Dataset & Human interaction and usability dataset from structured comparison tasks. & Knowledge organization \\

\cite{pangakis2025keeping} & CSS Annotation Datasets & Collection of 27 annotation tasks from computational social science studies. & Human-centered annotation \\

\cite{zhang2023human} & GPT-3 Schema Dataset, XPO Ontology & Human-curated schemas generated from GPT-3 outputs aligned with ontology knowledge. & Schema induction \\

\cite{deng2024reliable} & Mars Rover, ECLSS, KPI AIOps & Industrial and simulation time-series anomaly detection datasets. & Anomaly detection \\

\cite{bakhteev2025embed2discover} & Swiss Parliament Dataset & Parliamentary documents annotated for public broadcasting relevance. & Content analysis \\

\cite{bellos2025towards} & Human--AI Interaction Dataset & Multimodal recordings with task annotations and interaction analysis. & Human--AI collaboration \\

\cite{thielmann2024human} & 20 Newsgroups, BBC, M10 & Benchmark corpora for topic modeling and coherence evaluation. & Topic modeling \\

\cite{he2025enabling} & CDD, CUAD & Name screening and contract understanding datasets with simulated drift. & Self-improving agents \\

\cite{asiri2024towards} & PhishingArmy, PhishTank, OpenPhish & Phishing URL datasets with synthetic benign URL generation. & Phishing detection \\

\cite{bonet2023semi} & Annotated News Datasets & Reliable and unreliable news datasets for disinformation analysis. & Disinformation detection \\

\cite{schroeder2025just} & NYC/Boston Conversation Datasets & Community conversation datasets with thematic coding labels. & LLM-assisted annotation \\

\cite{wang2024rlhfpoison} & PKU-SafeRLHF, Alpaca & RLHF preference and instruction-following datasets for reward poisoning analysis. & RLHF security \\

\cite{liu2024aligning} & UltraFeedback, Anthropic HH & Preference-labeled dialogue datasets for alignment learning. & Human preference alignment \\

\cite{li2025iprop} & TEC, GROUNDED-EMOTIONS, TALES-EMOTION & Emotion classification datasets from tweets and stories. & Prompt optimization \\

\cite{chen2025instructioncp} & Multilingual Instruction Dataset & Chinese and Japanese instruction-following datasets for transfer learning. & Multilingual tuning \\

\cite{samani2025large} & EDOS Dataset & Hierarchical sexism detection benchmark dataset. & Explainable sexism detection \\

\cite{harada2025cognitive} & Zurich CLP Corpus & EEG-based sentiment reading dataset with cognitive annotations. & Cognitive feedback learning \\

\cite{martin2026beyond} & Crowdsourced Sentiment/Aesthetics Datasets & Sentiment and image aesthetics datasets with human interaction signals. & HITL learning \\

\cite{pusch2026human} & Movie KG, Hyena KG, MaRDI KG & Synthetic and real-world knowledge graph datasets for QA evaluation. & Knowledge graph QA \\

\cite{carvell2026human} & NATS Training Dataset & Air traffic control simulator and assessment datasets. & AI agent testing \\

\cite{wenz2025benchpress} & SPIDER, BIRD, FIBEN, BEAVER & Public and enterprise Text-to-SQL benchmark datasets. & Text-to-SQL \\

\cite{chhetri2025structsense} & Neuroscience Extraction Dataset & Scientific literature and assessment instruments for information extraction. & Structured extraction \\

\cite{chualost} & RabakBench & Multilingual safety benchmark with adversarial toxic prompts. & Multilingual safety \\

\cite{liu2025storm} & STORM-BORN Dataset & Mathematical derivation dataset curated from arXiv papers. & Mathematical reasoning \\

\cite{sadruddin2025llms4schemadiscovery} & Scientific Schema Corpus & Curated and uncurated scientific papers for schema discovery. & Schema mining \\

\cite{chu2025llm} & Pedagogical Response Dataset & Expert-graded short-answer educational responses. & Automated grading \\

\cite{schroeder2025large} & Virtual Character Review Dataset & Pilot and scoping review datasets with extracted research variables. & Systematic review extraction \\

\cite{hong2025dial} & Chinese Dialogue Dataset & Customer service dialogue dataset with 1,507 intent clusters. & Intent clustering \\

\cite{afzal2024towards} & SAP HR Dataset & HR FAQ, user utterance, and knowledge-base article datasets. & Retrieval QA \\

\cite{cohn2024towards} & C2STEM Corpus & Educational discourse corpus with transcripts and interaction logs. & Discourse analysis \\

\cite{zeng2024similar} & MIT Places365, AMLSim & Image and transaction datasets for contextual similarity analysis. & Similarity detection \\

\hline
\end{tabular}
}
\caption{Summary of datasets used in the reviewed human-in-the-loop and trustworthy NLP studies.}
\label{tab:dataset_summary}
\end{table*}

\end{document}